%% file: main.tex
\documentclass[11pt]{article}
\usepackage{eamt20}
\usepackage{times}
\usepackage{url}
\usepackage{latexsym}
\usepackage{enumitem}
\usepackage[small,bf]{caption} 
\usepackage{todonotes}
\usepackage{booktabs}
\usepackage{amsmath}
\usepackage{comment}
\usepackage[]{natbib}
\usepackage[official]{eurosym}
\usepackage[utf8]{inputenc}

\title{Correct Me If You Can: Learning from Error Corrections and Markings}

\author{Julia Kreutzer$^{\ast}$ \and Nathaniel Berger$^{\ast}$ \and Stefan Riezler$^{\dagger,\ast}$ \\
  $^{\ast}$Computational Linguistics \& $^\dagger$IWR\\
  Heidelberg University, Germany \\
  {\tt \{kreutzer, berger, riezler\}@cl.uni-heidelberg.de}\\
  }
\date{}

\begin{document}
\maketitle
\begin{abstract}
 Sequence-to-sequence learning involves a trade-off between signal strength and annotation cost of training data. For example, machine translation data range from costly expert-generated translations that enable supervised learning, to weak quality-judgment feedback that facilitate reinforcement learning. We present the first user study on annotation cost and machine learnability for the less popular annotation mode of error markings. We show that error markings for translations of TED talks from English to German allow precise credit assignment while requiring significantly less human effort than correcting/post-editing, and that error-marked data can be used successfully to fine-tune neural machine translation models.
\end{abstract}

\section{Introduction}
Successful machine learning for structured output prediction requires the effort of annotating sufficient amounts of gold-standard outputs---a task that can be costly if structures are complex and expert knowledge is required, as for example in neural machine translation (NMT) \citep{BahdanauETAL:15}. Approaches that propose to train sequence-to-sequence prediction models by reinforcement learning from task-specific scores, for example BLEU in machine translation (MT), shift the problem by simulating such scores by evaluating machine translation output against expert-generated reference structures \citep{RanzatoETAL:16, BahdanauETAL:17,  kreutzer-etal-2017-bandit, SokolovETAL:17}. 
An alternative approach that proposes to considerably reduce human  annotation effort by allowing to mark errors in machine outputs, for example erroneous words or phrases in a machine translation, has recently been proposed and been investigated in simulation studies by \citet{MarieMax:15, DomingoETAL:17, PetrushkovETAL:18}.
This approach takes the middle ground between supervised learning from error corrections as in machine translation post-editing\footnote{In the following we will use the more general term error corrections and MT specific term post-edits interchangeably.} (or from translations created from scratch) and reinforcement learning from sequence-level bandit feedback (this includes self-supervised learning where all outputs are rewarded uniformly). Error markings are highly promising since they suggest an interaction mode with low annotation cost, yet they can enable precise token-level credit/blame assignment, and thus can lead to an effective fine-grained discriminative signal for machine learning and data filtering.

Our work is the first to investigate learning from error markings in a user study. Error corrections and error markings are collected from junior professional translators, analyzed, and used as training data for fine-tuning neural machine translation systems. The focus of our work is on the learnability from error corrections and error markings, and on the behavior of annotators as teachers to a machine translation system. We find that error markings require significantly less effort (in terms of key-stroke-mouse-ratio (KSMR) and time) and result in a lower correction rate (ratio of words marked as incorrect or corrected in a post-edit). Furthermore, they are less prone to over-editing than error corrections.  
Perhaps surprisingly, agreement between annotators of which words to mark or to correct was lower for markings than for post-edits. However, despite of the low inter-annotator agreement, fine-tuning of neural machine translation could be conducted successfully from data annotated in either mode. 
Our data set of error corrections and markings is publicly available.\footnote{\url{https://www.cl.uni-heidelberg.de/statnlpgroup/humanmt/}}


\section{Related Work}\label{sec:related}
Prior work closest to ours is that of \citet{MarieMax:15,DomingoETAL:17,PetrushkovETAL:18}, however, these works were conducted by simulating error markings by heuristic matching of machine translations against independently created human reference translations. Thus the question of the practical feasibility of machine learning from noisy human error markings is left open.

User studies on machine learnability from human post-edits, together with thorough performance analyses with mixed effects models, have been presented by \citet{GreenETAL:14,BentivogliETAL:16,KarimovaETAL:18}. Albeit showcasing the potential of improving NMT through human corrections of machine-generated outputs, these works do not consider ``weaker'' annotation modes like error markings. User studies on the process and effort of machine translation post-editing are too numerous to list---a comprehensive overview is given in \citet{Koponen:16}. In contrast to works on interactive-predictive translation \citep{foster1997target, KnowlesKoehn:16, peris2017interactive, DomingoETAL:17, LamETAL:2018}, our approach does not require an online interaction with the human and allows to investigate, filter, pre-process, or augment the human feedback signal before making a machine learning update.

Machine learning from human feedback beyond the scope of translations, has considered learning from human pairwise preferences \citep{ChristianoETAL:17}, from human corrective feedback \citep{CeleminETAL:18}, or from sentence-level reward signals on a Likert scale \citep{KreutzerETAL:18}. However, none of these studies has considered error markings on tokens of output sequences, despite its general applicability to a wide range of learning tasks.

\section{User Study on Human Error Markings and Corrections}\label{sec:user}
The goal of the annotation study is to compare the novel error marking mode to the widely adopted machine translation post-editing mode. We are interested in finding an interaction scenario that costs little time and effort, but still allows to teach the machine how to improve its translations. In this section we present the setup, measure and compare the observed amount of effort and time that went into these annotations, and discuss the reliability and adoption of the new marking mode. Machine learnability, i.e. training of an NMT system on human-annotated data is discussed in Section~\ref{sec:nmt}.

\subsection{Participants}
We recruited $10$ participants that described themselves as native German speakers and having either a C1 or C2 level in English, as measured by the Common European Framework of Reference levels. $8$ participants were students studying translation or interpretation and $2$ participants were students studying computational linguistics. All participants were paid 100\euro{} for their participation in the study, which was done online, and limited to a maximum of 6 hours, and it took them between 2 and 4.5 hours excluding breaks. They agreed to the usage of the recorded data for research purposes.

\subsection{Interface}
\begin{figure}
    \centering
    \includegraphics[width=\columnwidth]{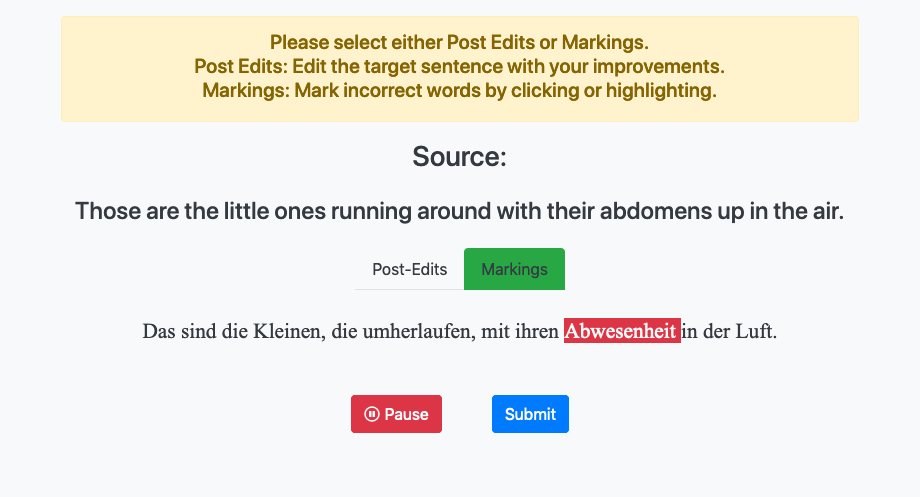}
    \caption{Interface for marking of translation outputs following user choice between markings and post-edits.}
    \label{fig:interface}
\end{figure}

The annotation interface has three modes: (1) markings, (2) corrections, and (3) the user-choice mode, where annotators first choose between (1) and (2) before submitting their annotation. While the first two modes are used for collecting training data for the MT model, the third mode is used for evaluative purposes to investigate which mode is preferable when given the choice. In any case, annotators are presented the source sentence, the target sentence and an instruction to either mark or correct (aka post-edit) the translation or choose an editing mode. They also had the option to pause and resume the session. No document-level context was presented, i.e., translated sentences were judged in isolation, but in consecutive order like they appeared in the original documents to provide a reasonable amount of context.
They received detailed instructions (see Appendix~\ref{sec:instructions}) on how to proceed with the annotation. Each annotator worked on 300 sentences, 100 for each mode, and an extra 15 sentences for intra-annotator agreement measures that were repeated after each mode. After the completion of the annotation task they answered a survey about the preferred mode, the perceived editing/marking speed, user-choice policies, and suggestions for improvement. A sreenshot of the interface showing a marking operation is shown in Figure \ref{fig:interface}.
The code for the interface is publicly available\footnote{\url{https://github.com/StatNLP/mt-correct-mark-interface}}.

\subsection{Data}
We selected a subset of $30$ TED talks to create the three data sets from the IWSLT17 machine translation corpus\footnote{\url{https://sites.google.com/site/iwsltevaluation2017/}}. The talks were filtered by the following criteria: single speakers, no music/singing, low intra-line final-sentence punctuation (indicating bad segmentation), length between 80 and 149 sentences. One additional short talk was selected for testing the inter- and intra-annotator reliability. We filtered out those sentences where model hypothesis and references were equal, in order to save annotation effort where it is clearly not needed, and also removed the last line from every talk (usually ``thank you''). For each talk, one topic of a set of keywords provided by TED was selected. 
See Appendix~\ref{sec:splits} for a description of how data was split across annotators.

\subsection{Effort and Time}
Correcting one translated sentence took on average approximately 5 times longer than marking errors, and required 42 more actions, i.e., clicks and keystrokes. That is 0.6 actions per character for post-edits, while only 0.03 actions per character for markings. This measurement aligns with the unanimous subjective impression of the participants that they were faster in marking mode. 

To investigate the sources of variance affecting time and effort, we use Linear Mixed Effect Models (LMEM) \citep{BarrETAL:13} and build one with KSMR as response variable, and another one for the total edit duration (excluding breaks) as response variable, and with the editing \texttt{mode} (correcting vs. marking) as fixed effect. 
For both response variables, we model users\footnote{Random effects are denoted, e.g., by $\texttt{(1|user\_id))}$.}, talks and target lengths\footnote{Target lengths measured by number of characters were binned into two groups at the limit of 176 characters.} as random effects, e.g., the one for KSMR:
\begin{align}
    \texttt{KSMR} \sim \texttt{mode} &+ (1 \mid \texttt{user\_id}) + (1 \mid \texttt{talk\_id}) \nonumber \\ 
    & + (1 \mid \texttt{trg\_length})
\end{align}

We use the implementation in the R package \texttt{lmer4} \citep{BatesETAL:15} and fit the models with restricted maximum likelihood. Inspecting the intercepts of the fitted models, we confirm that KSMR is significantly ($p=0.01$) higher for post edits than for markings (+3.76 on average). The variance due to the user (0.69) is larger than due to the talk (0.54) and the length (0.05)\footnote{Note that KSMR is already normalized by reference length, hence the small effect of target length. In a LMER for the raw action count (clicks+key strokes), this effect had a larger impact.}. Longer sentences have a slightly higher KSMR than shorter ones. When modeling the topics as random effects (rather than the talks), the highest KSMR (judging by individual intercepts) was obtained for physics and biodiversity and the lowest for language and diseases. This might be explained by e.g. the MT training data or the raters expertise.

Analyzing the LMEM for editing duration, we find that post-editing takes on average 42s longer than marking, which is significant at $p=0.01$. The variance due to the target length is the largest, followed by the one due to the talk and the one due to the user is smallest.
Long sentences have a six time higher editing duration on average than shorter ones. With respect to topics, the longest editing was done for topics like physics and evolution, shortest for diseases and health. 

\subsubsection{Annotation Quality}\label{sec:quality}
The corrections increased the quality, measured by comparison to reference translations, by 2.1 points in BLEU and decreased TER by 1 point. While this indicates a general improvement, it has to be taken with a grain of salt, since the post-edits are heavily biased by the structure, word choice etc. by the machine translation, which might not necessarily agree with the reference translations, while still being accurate.

\begin{table}[t]
    \centering
    \resizebox{\columnwidth}{!}{%
    \begin{tabular}{l|l}
        \toprule
        src & I am a nomadic artist. \\
        hyp & Ich bin ein nomadischer Künstler. \\
        pe & Ich bin ein nomadischer Künstler. \\
        ref & Ich \underline{wurde} zu einer nomadischen Künstler\underline{in}. \\
        \midrule
        src &           I look at the chemistry of the ocean today.  \\
        hyp & Ich betrachte \underline{heute} die Chemie des Ozeans. \\
        pe & Ich erforsche \underline{täglich} die Chemie der Meere. \\
        ref & Ich untersuche die Chemie der Meere \underline{der Gegenwart}.\\
        \midrule
        src & There's even a software called \underline{cadnano} that allow \dots \\
        hyp & Es gibt sogar eine Software namens \underline{Caboano}, die \dots \\
        pe & Es gibt sogar eine Software namens \underline{Caboano}, die \dots \\
        ref & Es gibt sogar eine Software namens \underline{"cadnano"}, \dots \\
        \midrule
        src & It was a thick forest. \\
        hyp & Es \underline{war} ein dicker Wald.\\
        pe & Es \underline{handelte sich um} einen dichten Wald.\\
        ref &  \underline{Auf der Insel war} dichter Wald.\\
        \bottomrule
    \end{tabular}
    }
    \caption{Examples of post-editing to illustrate differences between reference translations (\emph{ref}) and post-edits (\emph{pe}). Example 1: The gender in the German translation could not be inferred from the context, since speaker information is unavailable to post-editor. Example 2: ``today'' is interpreted as adverb by the NMT, this interpretation is kept in the post-edit (``telephone game'' effect). Example 3: Another case of the ``telephone game'' effect: the name of the software is changed by the NMT, and not corrected by post-editors. Example 4: Over-editing by post-editor, and more information in the reference translation than in the source.}
    \label{tab:examples-pe}
\end{table}

\paragraph{How good are the corrections?}
We therefore manually inspect the post-edits to get insights into the differences between post-edits and references. Table~\ref{tab:examples-pe} provides a set of examples\footnote{Selected because of their differences to references.} with their analysis in the caption. Besides the effect of ``literalness'' \citep{Koponen:16}, we observe three major problems:
\begin{enumerate}
    \item \emph{Over-editing}: Editors edited translations even though they are adequate and fluent.
    \item \emph{``Telephone game'' effect}: Semantic mistakes (that do not influence fluency) introduced by the MT system flow into the post-edit and remain uncorrected, when more obvious corrections are needed elsewhere in the sentence.
    \item \emph{Missing information}: Since editors only observe a portion of the complete context, i.e., they do not see the video recording of the speaker or the full transcript of the talk, they are not able to convey as much information as the reference translations.
\end{enumerate}

\begin{table}[t]
    \centering
    \resizebox{\columnwidth}{!}{%
    \begin{tabular}{l|l}
    \toprule
    src & Each year, it sends up a new generation  of shoots. \\
    ann & Jedes Jahr \underline{sendet} es eine neue Generation von \underline{Shoots.} \\
    sim &  Jedes Jahr \underline{sendet es eine} neue \underline{Generation von Shoots}. \\
    ref & Jedes Jahr wachsen neue Triebe.  \\
    \midrule
    src &  He killed 63 percent of the Hazara population. \\
    ann & Er \underline{starb} 63 Prozent der Bevölkerung \underline{Hazara.} \\
    sim & Er \underline{starb} 63 \underline{Prozent} der \underline{Bevölkerung} \underline{Hazara.} \\
    ref & Er tötete 63\% der Hazara-Bevölkerung. \\
    \midrule
    src & They would ordinarily support fish and other wildlife. \\
    ann & Sie \underline{würden} Fisch und andere wild lebende Tiere unterstützen. \\
    sim & \underline{Sie} würden Fisch und andere \underline{wild} \underline{lebende} \underline{Tiere} \underline{unterstützen}. \\ 
    ref & Normalerweise würden sie Fisch und andere Wildtiere ernähren. \\ 
    \bottomrule
    \end{tabular}
    }
    \caption{Examples of markings to illustrate differences between human markings (\emph{ann}) and simulated markings (\emph{sim}). Marked parts are underlined. Example 1: ``es'' not clear from context, less literal reference translation. Example 2: Word omission (preposition after ``Bevölkerung'') or incorrect word order is not possible to mark. Example 3: Word order differs between MT and references, word omission (``ordinarily'') not marked.}
    \label{tab:examples-markings}
\end{table}

\paragraph{How good are the markings?}
Markings, in contrast, are less prone to over-editing, since they have fewer degrees of freedom. They are equally exposed to problem (3) of missing context, and another limitation is added: Word omissions and word order problems cannot be annotated. Table~\ref{tab:examples-markings} gives a set of examples that illustrate these problems. While annotators were most likely not aware of problems (1) and (2), they might have sensed that information was missing, as well as the additional limitations of markings. The simulation of markings from references as used in previous work \citep{PetrushkovETAL:18, MarieMax:15} seems overly harsh for the generated target translations, e.g., marking ``Hazara-Bevölkerung'' as incorrect, even though it is a valid translation of ``Hazara population''. 

\begin{table}[h]
    \centering
    \resizebox{\columnwidth}{!}{%
    \begin{tabular}{l|c|c}
        \toprule
        \textbf{Mode} & \textbf{Intra-Rater (Mean / Std.)} $\alpha$ &\textbf{ Inter-Rater} $\alpha$ \\
        \midrule
        Marking & $0.522$ / $0.284$ & $0.201$ \\
        Correction & $0.820$ / $0.171$ & $0.542$ \\
        User-Chosen & $0.775$ / $0.179$ & $0.473$ \\
        \bottomrule
    \end{tabular}
    }
    \caption{Intra- and Inter-rater agreement calculated by Krippendorff's $\alpha$.}
    \label{tab:agreement}
\end{table}

\paragraph{How reliable are corrections and markings?}
In addition to the absolute quality of the annotations, we are interested in measuring their reliability: Do annotators agree on which parts of a translation to mark or edit? While there are many possible valid translations, and hence many ways to annotate one given translation, it has been shown that learnability profits from annotations with less conflicting information \citep{KreutzerETAL:18}. In order to quantify agreement for both modes on the same scale, we reduce both annotations to sentence-level quality judgments, which for markings is the ratio of words that were marked as incorrect in a sentence, and for corrections the ratio of words that was actually edited. If the hypothesis was perfect, no markings nor edits would be required, and if it was completely wrong, all of it had to be marked or edited. After this reduction, we measure agreement with Krippendorff's $\alpha$ \citep{Krippendorff:13}, see Table \ref{tab:agreement}.

\paragraph{Which mode do annotators prefer?}
In the user-choice mode, where annotators can choose for each sentence whether they would like to mark or correct it, markings were chosen much more frequently than post-edits (61.9\%). Annotators did not agree on the preferred choice of mode for the repeated sentences ($\alpha=-0.008$), which indicates that there is no obvious policy when one of the modes would be advantageous over the other. In the post-annotation questionnaire, however, 60\% of the participants said they generally preferred post-edits over markings, despite markings being faster, and hence resulting in a higher hourly pay. 

To better understand the differences in modes, we asked them about their policies in the user-choice mode where for each sentence they would have to decide individually if they want to mark or post-edit it. The most commonly described policy is decide based on error types and frequency: choose post-edits when insertions or re-ordering is needed, and markings preferably for translations with word errors (less effort than doing a lookup or replacement). One person preferred post-edits for short translations, markings for longer ones, another three generally preferred markings generally, and one person preferred post-edits.
Where annotators found the interface to need improvements was (1) in the presentation of inter-sentential context, (2) in the display of overall progress and (3) an option to edit previously edited sentences. For the marking mode they requested an option to mark missing parts or areas for re-ordering.

\begin{figure}
    \centering
    \includegraphics[width=\columnwidth]{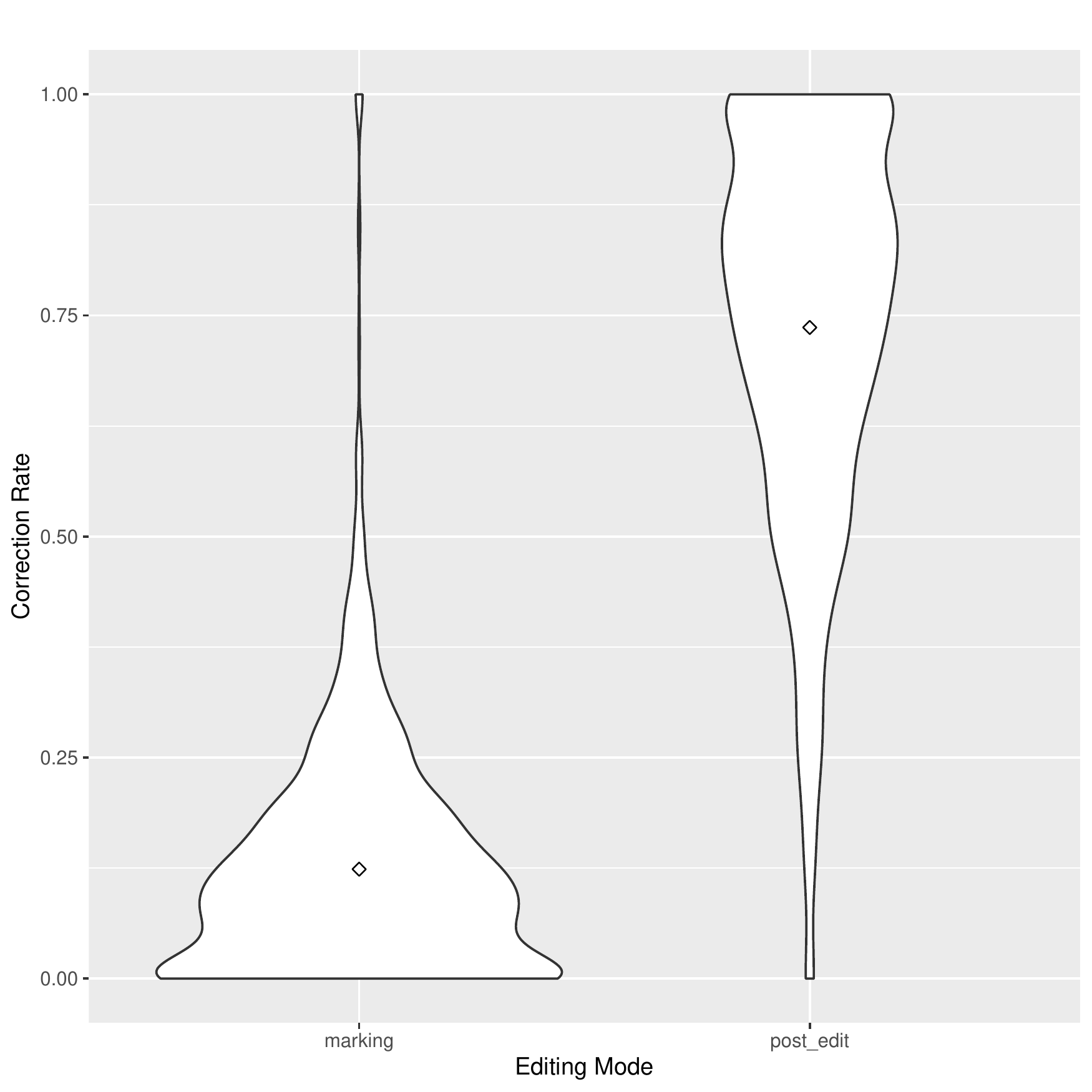}
    \caption{Correction rate by annotation mode. The correction rate describes the ratio of words in the translation that were marked as incorrect (in marking mode) or edited (in post-editing mode). Means are indicated with diamonds.}
    \label{fig:violin}
\end{figure}
\paragraph{Do markings and corrections express the same translation quality judgment?}
 We observe that annotators find more than twice as many token corrections in post-edit mode than in marking mode\footnote{The automatically assessed translation quality for the baseline model does not differ drastically between the portions selected per mode.} 

This is partially caused by the reduced degrees of freedom in marking mode, but also underlines the general trend towards over-editing when in post-edit mode. If markings and post-edits were used to compute a quality metric based on the correction rate, translations are judged as much worse in post-editing mode than in marking mode (Figure~\ref{fig:violin}). This also holds for whole sentences, where 273 (26.20\%) were left un-edited in marking mode, and only 3 (0.29\%) in post-editing mode.

\section{Machine Learnability of NMT from Human Markings and Corrections}\label{sec:nmt}
The hypotheses presented to the annotators were generated by an NMT model. The goal is to use the supervision signal provided by the human annotation to improve the underlying model by machine learning. Learnability is concerned with the question of how strong a signal is necessary in order to see improvements in NMT fine-tuning on the respective data. 

\paragraph{Definition.}
Let $x = x_1 \dots x_S$ be a sequence of indices over a source vocabulary $\mathcal{V}_{\textsc{Src}}$, and $y = y_1 \dots y_T$ a sequence of indices over a target vocabulary $\mathcal{V}_{\textsc{Trg}}$. The goal of sequence-to-sequence learning is to learn a function for mapping a input sequence $x$  into an output sequences $y$. For the example of machine translation, $y$ is a translation of $x$, and a model parameterized by a set of weights $\theta$ is optimized to maximize $p_{\theta}(y \mid x)$. This quantity is further factorized into conditional probabilities over single tokens 
$p_{\theta}(y \mid x) = \prod^{T}_{t=1}{p_{\theta}(y_t \mid x; y_{<t})}$, 
where the latter distribution is defined by the neural model's softmax-normalized output vector:
\begin{align}\label{eq:softmax}
    p_{\theta}(y_t \mid x; y_{<t}) = \texttt{softmax}(\text{NN}_{\theta}(x; y_{<t})).
\end{align}
There are various options for building the architecture of the neural model $\text{NN}_{\theta}$, such as recurrent \citep{BahdanauETAL:15}, convolutional \citep{GehringETAL:17} or attentional \citep{VaswaniETAL:17} encoder-decoder architectures (or a mix thereof \citep{ChenETALgoogle:18}).

\paragraph{Learning from Error Corrections.} The standard supervised learning mode in human-in-the-loop machine translation assumes a fully corrected output $y^*$ for an input $x$ that is treated similar to a gold standard reference translation \citep{TurchiETAL:17}. Model adaptation can be performed by maximizing the likelihood of the user-provided corrections where 
\begin{align}\label{eq:loss}
    L(\theta) = \sum_{x,y^*} \sum_{t=1}^T \log p_{\theta}(y^*_t \mid x; y^*_{<t}),
\end{align}
 using stochastic gradient descent techniques \citep{BottouETAL:18}.

\paragraph{Learning from Error Markings.} A weaker feedback mode is to let a human teacher mark the correct parts of the machine-generated output $\hat{y}$ \citep{MarieMax:15,PetrushkovETAL:18,DomingoETAL:17}. As a consequence every token in the output receives a reward $\delta^m_t$, either $\delta^+_t$ if marked as correct, or $\delta^-_t$ otherwise. \citet{PetrushkovETAL:18} proposed a model with $\delta^+_t=1$ and $\delta^-_t=0$, but this weighting schemes leads to the ignorance of incorrect outputs in the gradient and the rewarding of correct tokens. Instead, we find it beneficial to penalize incorrect tokens, with e.g. $\delta^-_t=-0.5$, and reward correct tokens $\delta^+_t=0.5$, which aligns with the findings from \citet{lam2019}. The objective of the learning system is to maximize the likelihood of the correct parts of the output where 
\begin{align} \label{eq:weighted-loss}
  L(\theta) = \sum_{x,\hat{y}} \sum_{t=1}^T \delta^m_t \log p_{\theta}(\hat{y}_t \mid x; \hat{y}_{<t}).
\end{align}

\subsection{NMT Fine-Tuning}

\begin{table}[]
    \centering
     \resizebox{\columnwidth}{!}{%
    \begin{tabular}{l|lll}
        \toprule
         \textbf{Domain} & \textbf{train} & \textbf{dev} & \textbf{test} \\
         \midrule
         WMT17 & 5,919,142 & \textcolor{gray}{2,169} & \textcolor{gray}{3,004} \\
         \textcolor{gray}{IWSLT17} & \textcolor{gray}{206,112} & \textcolor{gray}{2,385} & \textcolor{gray}{1,138}\\
         Selection &  1035 corr / 1042 mark & & 1,043\\
         \bottomrule
    \end{tabular}
    }
    \caption{Data sizes (en-de), official splits from WMT17 and IWSLT17. Our target-domain data is a subset of selected talks from IWSLT2017 training data totalling 3,120 sentences.}
    \label{tab:data}
\end{table}
\paragraph{NMT Model and Data.}
The goal is to adapt a general-domain NMT model to a new domain with either post-edits or markings. For the general-domain NMT system, we use the pre-trained 4-layer LSTM encoder-decoder Joey NMT WMT17 model \citep{kreutzer-etal-2019-joey} for translations from English to German\footnote{Pre-trained model: \url{https://github.com/joeynmt/joeynmt/blob/master/README.md#wmt17}; modified fork of Joey NMT: \url{https://github.com/StatNLP/joeynmt/tree/mark}}. The model is trained on a joint vocabulary with 30k subwords \citep{sennrich-etal-2016-neural}. Model outputs are de-tokenized and un-BPEd before being presented to the annotators. 
With the help of human annotations we then adapt this model to the domain of TED talk transcripts by continuing learning on the annotated data. Hyperparameters including learning rate schedule, dropout and batch size for this fine-tuning step are tuned on the IWSLT17 dev set. For the marking mode, the weights $\delta^+$ and $\delta^-$ are tuned in addition.
As test data, we use the split of the selected talks that was annotated in the user-mode, since the purpose of this split was the evaluation of user preference. There is no overlap in the three data splits, but they have the same distribution over topics, so that we can both measure local adaptation and draw comparisons between modes. Data sizes are given in Table~\ref{tab:data}.

\paragraph{Evaluation.}
The models are evaluated with TER \citep{snover2006study}, BLEU \citep{papineni-etal-2002-bleu} and METEOR \citep{LavieETAL:09}\footnote{Computed with MultEval v0.5.1 \citep{Clark:11} on tokenized outputs.} against references translations. Significance is tested with approximate randomization for three runs for each system \citep{Clark:11}.

\subsection{Results}
\input{tab-real}
\paragraph{Corrections, Markings and Quality Judgments.}
Table~\ref{tab:real} compares the models after fine-tuning with corrections and markings with the original WMT out-of-domain model.

The ``small'' model trained with error corrections is trained on one fifth of the data, which is comparable to the effort it takes to collect the error markings. Both error corrections and markings can be reduced to sentence-level quality judgments, where all tokens receive the same weight in Eq.~\label{eq:weighted-loss} $\delta = \frac{\# marked}{hyp tokens}$ or $\delta = \frac{\# corrected}{hyp tokens}$. In addition, we compare the markings against a random choice of marked tokens per sentence.\footnote{Each token is marked with probability $p_{mark} = 0.5$.} We see that both models trained on corrections and markings improve significantly over the baseline (rows 2 and 3). Tuning the weights for (in)correct tokens makes a small but significant difference for learning from markings (rows 4 and 5). These human markings lead to significantly better models than random markings (row 6). When reducing both types of human feedback to sentence-level quality judgments, no loss in comparison to error corrections and a small loss for markings (rows 7 and 8) is observed. We suspect that the small margin between results for learning from corrections and markings is due to evaluating against references. Effects like over-editing (see Section~\ref{sec:quality}) produce training data that lead the model to generate outputs that diverge more from independent references and therefore score lower than deserved under all metrics except for METEOR. 

\paragraph{Human Evaluation.}
It is infeasible to collect markings or corrections for all our systems for a more appropriate comparison than to references, but for that purpose we conduct a small human evaluation study. Three bilingual raters receive 120 translations of the test set ($\sim$10\%) and the corresponding source sentences for each mode  and judge whether the translation is better, as good as, or worse than the baseline: 64\% of the translations obtained from learning from error markings are judged at least as good as the baseline, compared to 65.2\% for the translations obtained from learning from error corrections. Table~\ref{tab:human-eval} shows the detailed proportions excluding identical translations.

\begin{table}[h]
    \centering
    \begin{tabular}{l|ccc}
        \toprule
        \textbf{System} & $>$ BL & $=$ BL  & $<$ BL \\
        \midrule
        Error Markings & \textbf{43.0\%} & 21.0\% & 36.4\% \\
        Error Corrections & \textbf{49.1\%} & 16.1\% & 34.7\%\\
        \bottomrule
    \end{tabular}
    \caption{Human preferences for comparisons between baseline (BL) translations and the NMT system fine-tuned on error markings and corrections. $>$: better than the baseline, $<$ worse than the baseline.} 
    \label{tab:human-eval}
\end{table}

\begin{figure}
    \centering
  \includegraphics[width=\columnwidth]{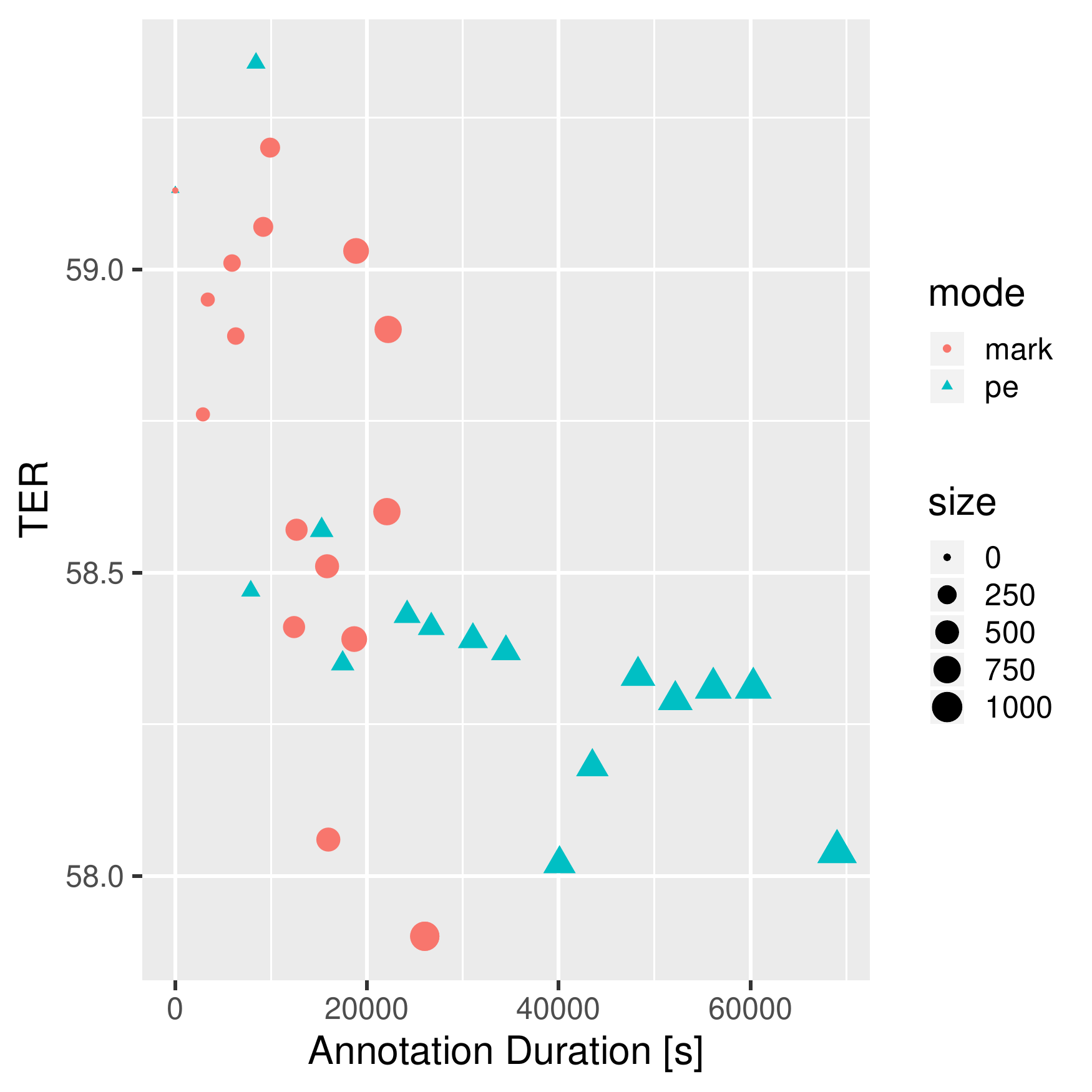}
    \caption{Improvement in TER for training data of varying size: lower is better. Scores are collected across two runs with a random selection of $k \in [125, 250, 375, 500, 625, 750, 875]$ training data points.}
    \label{fig:train_size}
\end{figure}

\paragraph{Effort vs. Translation Quality.}
Figure~\ref{fig:train_size} illustrates the relation between the total time spent on annotations and the resulting translation quality for corrections and markings trained on a selection of subsets of the full annotated data: The overall trend shows that both modes benefit from more training data, with more variance for the marking mode, but also a steeper descent. From a total annotation amount of approximately 20,000s on ($\approx$ 5.5h), markings are the more efficient choice.

\subsubsection{LMEM Analysis}
We fit a LMEM for sentence-level quality scores of the baseline, and three runs each for the NMT systems fine-tuned on markings and post-edits respectively, and inspect the influence of the system as a fixed effect, and sentence id, topic and source length as random effects.
 \begin{align}
     \texttt{TER} \sim& \texttt{system} + (1 \mid \texttt{talk\_id}/\texttt{sent\_id}) \nonumber\\
     &+ (1 \mid \texttt{topic}) + (1 \mid \texttt{src\_length})  \nonumber
 \end{align}
The fixed effect is significant at $p=0.05$, i.e., the quality scores of the three systems differ significantly under this model. The global intercept lies at 64.73, the one for marking 1.23 below, and the one for post-editing 0.96 below. The variance in TER is for the largest part explained by the sentence, then the talk, the source length, and the least by the topic.

\section{Conclusion}
We presented the first user study on the annotation process and the machine learnability of human error markings of translation outputs. This annotation mode has so far been given less attention than error corrections or quality judgments, and has until now only been investigated in simulation studies. We found that both according to automatic evaluation metrics and by human evaluation,  fine-tuning of NMT models achieved comparable gains by learning from error corrections and markings. However, error markings required several orders of magnitude less human annotation effort. \\
In future work we will investigate the integration of automatic markings into the learning process, and we will explore online adaptation possibilities.

\section*{Acknowledgments}

We would like to thank the anonymous reviewers for their feedback, Michael Staniek and Michael Hagmann for the help with data processing and analysis, and Sariya Karimova and Tsz Kin Lam for their contribution to a preliminary study. 
The research reported in this paper was supported in part by the German research foundation (DFG) under grant RI-2221/4-1.

\bibliography{acl2020}
\bibliographystyle{apalike}

\appendix

\section*{Appendix}
\section{Annotator Instructions}\label{sec:instructions}
The annotators received the following instructions:
\begin{itemize}[noitemsep,topsep=0pt]
\item You will be shown a source sentence, its translation and an instruction.
\item Read the source sentence and the translation.
\item Follow the instruction by either marking the incorrect words of the translation by clicking on them or highlighting them, correcting the translation by deleting, inserting and replacing words or parts of words, or choosing between modes (i) and (ii), and then click ``submit''.
    \begin{itemize}[noitemsep,topsep=0pt]
        \item In (ii), if you make a mistake and want to start over, you can click on the button ``reset''. 
        \item In (i), to highlight, click on the word you would like to start highlighting from, keep the mouse button pushed down, drag the pointer to the word you would like to stop highlighting on, and release the mouse button while over that word. 
    \end{itemize}
    \item If you want to take a short break (get a coffee, etc.), click on ``pause'' to pause the session. We're measuring time it takes to work on each sentence, so please do not overuse this button (e.g. do not press pause while you're making your decisions), but also do not feel rushed if you feel uncertain about a sentence.
    \item Instead, if you want to take a longer break, just log out. The website will return you return you to the latest unannotated sentence when you log back in. If you log out in the middle of an annotation, your markings or post-edits will not be saved.
    \item After completing all sentences (ca. 300), you'll be asked to fill a survey about your experience.
    \item Important:
\begin{itemize}[noitemsep,topsep=0pt]
    \item Please do not use any external dictionaries or translation tools.
    \item You might notice that some sentences re-appear, which is desired. Please try to be consistent with repeated sentences.
    \item There is no way to return and re-edit previous sentences, so please make sure you're confident with the edits/markings you provided before you click ``submit''.
\end{itemize}
\end{itemize}

\section{Creating Data Splits}\label{sec:splits}
In order to have users see a wider range of talks, each talk was split into three parts (beginning, middle, and end). Each talk part was assigned an annotation mode. Parts were then assigned to users using the following constraints:
\begin{itemize} [noitemsep,topsep=0pt]
    \item Each user should see nine document parts.
    \item No user should see the same document twice.
    \item Each user should see three sections in post-editing, marking, and user-choice mode. 
    \item Each user should see three beginning, three middle, and three ending sections.
    \item Each document should be assigned each of the three annotation modes.
\end{itemize}
To avoid assigning post-editing to every beginning section, marking to every middle section, and user-choice to every ending section, assignment was done with an integer linear program with the above constraints. Data was presented to users in the order [Post-edit, Marking, User Chosen, Agreement].

\end{document}

%% file: tab-real.tex
\begin{table}[ht]
    \centering
    \resizebox{\columnwidth}{!}{%
     \begin{tabular}{llccc}
         \toprule
         & \textbf{System }& \textbf{TER $\downarrow$} & \textbf{BLEU $\uparrow$} &\textbf{ METEOR $\uparrow$} \\
         \midrule
        1 & WMT baseline  & 58.6 & 23.9 & 42.7\\
         \midrule
         & \multicolumn{4}{l}{\textbf{Error Corrections}}\\
         \midrule
        2 & Full & 57.4$^\star$ & 24.6$^\star$ & 44.7$^\star$ \\
        3 & Small & 57.9$^\star$ & 24.1 & 44.2$^\star$ \\
         \midrule
         \multicolumn{4}{l}{\textbf{Error Markings}}\\
         \midrule

        4 & 0/1 & 57.5$^\star$ & 24.4$^\star$ & 44.0$^\star$ \\
        5 & -0.5/0.5 & 57.4$^\star$ & 24.6$^\star$ & 44.2$^\star$ \\
        6 & random & 58.1$^\star$ & 24.1 & 43.5$^\star$ \\
         \midrule
         \multicolumn{4}{l}{\textbf{Quality Judgments}} \\
         \midrule
        7 & from corrections & 57.4$^\star$ & 24.6$^\star$ & 44.7$^\star$ \\
        8 & from markings & 57.6$^\star$ & 24.5$^\star$ & 43.8$^\star$ \\
         \bottomrule
    \end{tabular}
   }
    \caption{Results on the test set with feedback collected from humans. Decoding with beam search of width 5 and length penalty of 1. Significant ($p<=0.05$) improvements over the baseline are marked with $^\star$. Full error corrections and error markings only significantly differ in terms of METEOR.}
    \label{tab:real}
\end{table}

